\documentclass[journal]{IEEEtran}
\RequirePackage{graphicx,xcolor}
 \usepackage{amsmath}
\ifCLASSINFOpdf
\else
\fi

\begin{document}
\title{At First Contact: Stiffness Estimation Using Vibrational Information for Prosthetic Grasp Modulation}

\author{Anway S. Pimpalkar,
        Ariel Slepyan,
        and Nitish V. Thakor
\thanks{A. S. Pimpalkar and N. V. Thakor are with the Department of Biomedical Engineering, Johns Hopkins University,  Baltimore, MD 21218, USA.}
\thanks{A. Slepyan and N. V. Thakor are with the Department of Electrical and Computer Engineering, Johns Hopkins University,  Baltimore, MD 21218, USA.}
\thanks{Corresponding author: A. Pimpalkar (e-mail: apimpal1@jh.edu)}
}

\maketitle

\begin{abstract}
Stiffness estimation is crucial for delicate object manipulation in robotic and prosthetic hands but remains challenging due to dependence on force and displacement measurement and real-time sensory integration. This study presents a piezoelectric sensing framework for stiffness estimation at first contact during pinch grasps, addressing the limitations of traditional force-based methods. Inspired by human skin, a multimodal tactile sensor that captures vibrational and force data is developed and integrated into a prosthetic hand's fingertip. Machine learning models, including support vector machines and convolutional neural networks, demonstrate that vibrational signals within the critical 15 ms after first contact reliably encode stiffness, achieving classification accuracies up to 98.6\% and regression errors as low as 2.39 Shore A on real-world objects of varying stiffness. Inference times of less than 1.5 ms are significantly faster than the average grasp closure time (16.65 ms in our dataset), enabling real-time stiffness estimation before the object is fully grasped. By leveraging the transient asymmetry in grasp dynamics, where one finger contacts the object before the others, this method enables early grasp modulation, enhancing safety and intuitiveness in prosthetic hands while offering broad applications in robotics.
\end{abstract}


\IEEEpeerreviewmaketitle

\section{Introduction}

\IEEEPARstart{T}ACTILE sensing is fundamental to human perception, enabling fine interactions with the environment and supporting dexterous motor control \cite{Johansson2009}. Through the integration of exteroceptive cues, the human somatosensory system allows real-time adjustments during object manipulation, ensuring safe and effective interactions. Replicating this capability in artificial systems such as robotics and prosthetics remains a significant challenge, particularly for tasks requiring precise stiffness estimation \cite{Stabile2022}.

Estimating object stiffness is crucial for applications requiring delicate manipulation. Traditional approaches predominantly rely on tactile force sensors or measuring hand displacements under applied forces \cite{Mahboubi2018, Andrecioli2012, Deng2020, Zhang2023}. In these methods, contact stiffness is typically inferred from the motor current and angular displacement of the prosthetic hand or as the ratio of force to deformation. These approaches become effective only after full contact has been established \cite{Zhang2023}, which increases the risk of damaging brittle objects in delicate tasks.

This study develops and validates a novel method for stiffness estimation at first contact using piezoelectric sensing, focusing on the previously unexplored period between the first and second finger contacts during a grasp. We demonstrate this approach in pinch grasps, a ubiquitous action in daily manipulation tasks \cite{Vergara2014}. Our dataset comprises silicone blocks with known stiffness properties, supplemented by validation on everyday objects. To underscore the value of this approach, we evaluate two machine learning methods for stiffness discrimination and regression, showing that the information encoded by the sensors effectively reflects object properties.

Critically, we demonstrate that tactile force sensors fail to capture meaningful data during the transient first contact phase, where vibrational signals are most informative for stiffness estimation. As prosthetics advance in processing and manipulation speed \cite{Billard2019}, integrating piezoelectric sensors to exploit this brief yet critical window addresses traditional limitations, enhancing adaptive control and ensuring safe interaction with fragile objects.

\section{Biomimetic Tactile Sensor Design}

\begin{figure}
    \centering    
    \includegraphics[width=\linewidth]{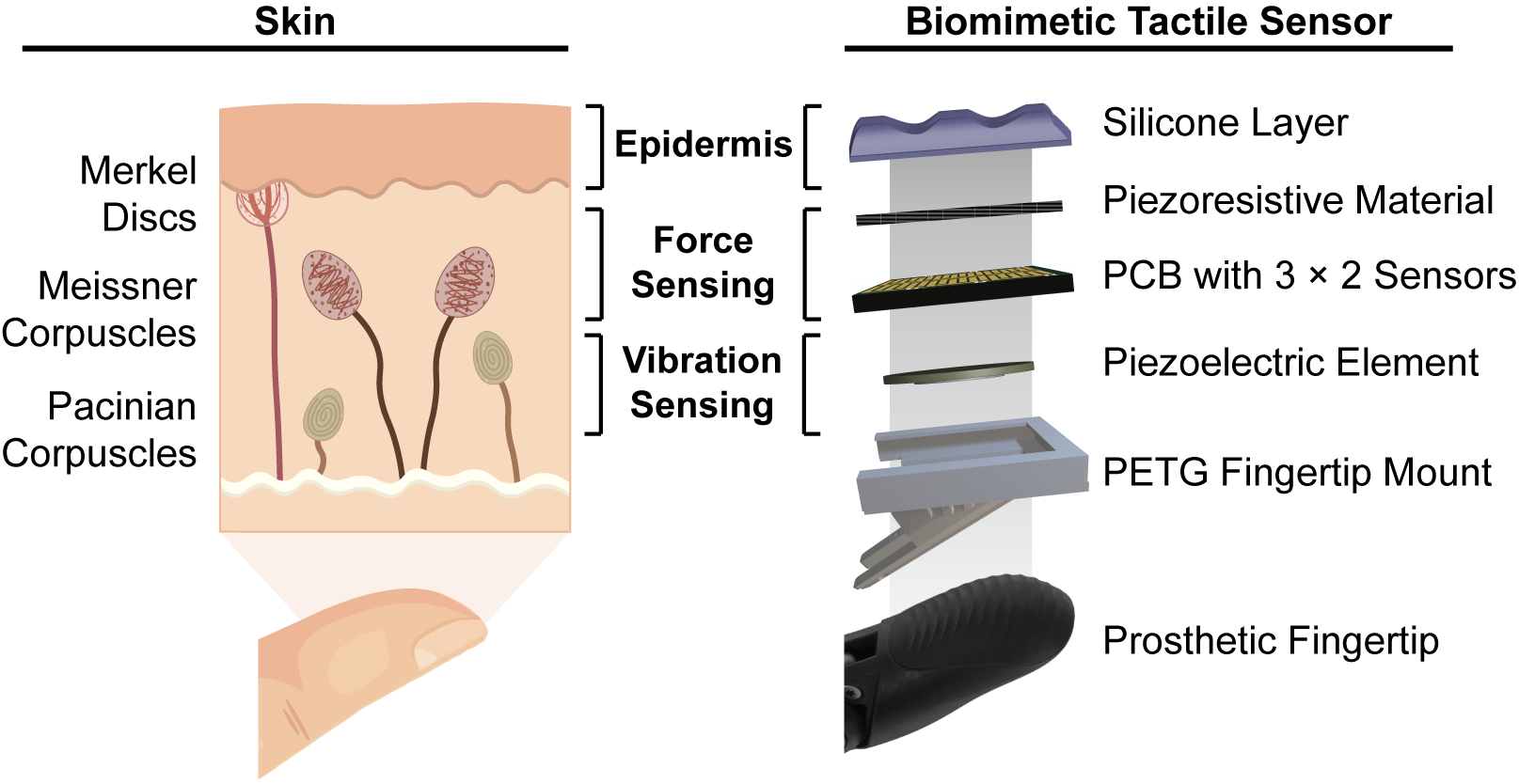}
    \caption{Multilayer biomimetic tactile fingertip sensor for vibration and force sensing, inspired by mechanoreceptor placement in human skin.}
    \label{fig:SensorDesign}
\end{figure}

We developed a biomimetic multilayer tactile sensor to capture both force and vibration information, mimicking the mechanoreceptor distribution in human skin \cite{Abraira2013}. We integrated it into the distal phalanges of the index finger and thumb on the TASKA HandGen2 (TASKA Prosthetics, New Zealand), as shown in Fig. \ref{fig:SensorDesign}. Each sensor assembly features grip tips with a 3 mm-thick molded silicone layer of 65 Shore A stiffness. The layer has dimpled extrusions on its underside, directing applied forces precisely onto the underlying force sensors. Beneath the silicone layer, a $20 \, \text{mm} \times 15 \, \text{mm}$ printed circuit board (PCB) is designed to house a $3 \times 2$ piezoresistive force sensor array on its front, mimicking the superficial placement of human skin receptors responsible for encoding sustained force information (Merkel discs and Meissner corpuscles). The sensor array, analogous to an orthogonal force sensor design, features interwoven traces serving as the two poles for the piezoresistive Velostat material layered on top. This 0.1 mm-thick sheet exhibits a volume resistivity of $<500 , \Omega$-cm and a surface resistivity of $<31,000$ $\Omega/\text{cm}^2$. For vibration sensing, a 10 mm diameter PZT piezoelectric element is mounted on the reverse side of the PCB, emulating the deeper placement of skin receptors responsible for encoding transient vibration information (Pacinian corpuscles). The assembly is encased in a custom 3D-printed polyethylene terephthalate glycol (PETG) fingertip mount, specifically designed for the prosthetic hand and secured with screws.

The vibration and force sensor outputs are buffered and amplified using an analog circuit for signal processing. Data acquisition is facilitated by a 10-bit resolution analog-to-digital converter integrated into a Teensy 4.0 MCU (ARM Cortex-M7, NXP iMXRT1062) operating at a clock speed of 600 MHz. Signals are recorded at an average sampling rate of 4936 Hz. The six piezoresistive sensor signals are time-division multiplexed using a row-column addressing scheme and measured via a voltage divider circuit. The analog output from the piezoelectric element is buffered through a non-inverting amplifier and offset to 1.65 V, enabling the detection of both positive and negative spikes within a 3.3 V range. Once buffered into an array, the data is transmitted to a connected computer over the serial port, where it is managed by a Python 3.8 script for further processing.

The vibration data undergoes exponential smoothing ($\alpha=0.5$) in real-time on the microcontroller. Force data is smoothed using a sliding window average. After the data is collected, a Savitzky-Golay filter is employed to further eliminate high-frequency noise without distorting key vibrational patterns.

\section{Stiffness Estimation at First Contact}
\subsection{Training Data Collection}
The experimental setup for stiffness estimation involved pinching silicone blocks to collect training data. Five blocks with dimensions of 7$\times$7$\times$3.5 cm were fabricated, representing stiffness values of 10, 20, 29, 43, and 60 Shore A, verified with a PCE-DD-A durometer (PCE Instruments, Germany). Desired stiffness levels were achieved by proportionally blending Smooth-On Mold Max 60 and Mold Max 10T (Smooth-On Inc., USA) silicone compounds.

\begin{figure}
    \centering    
    \includegraphics[width=\linewidth]{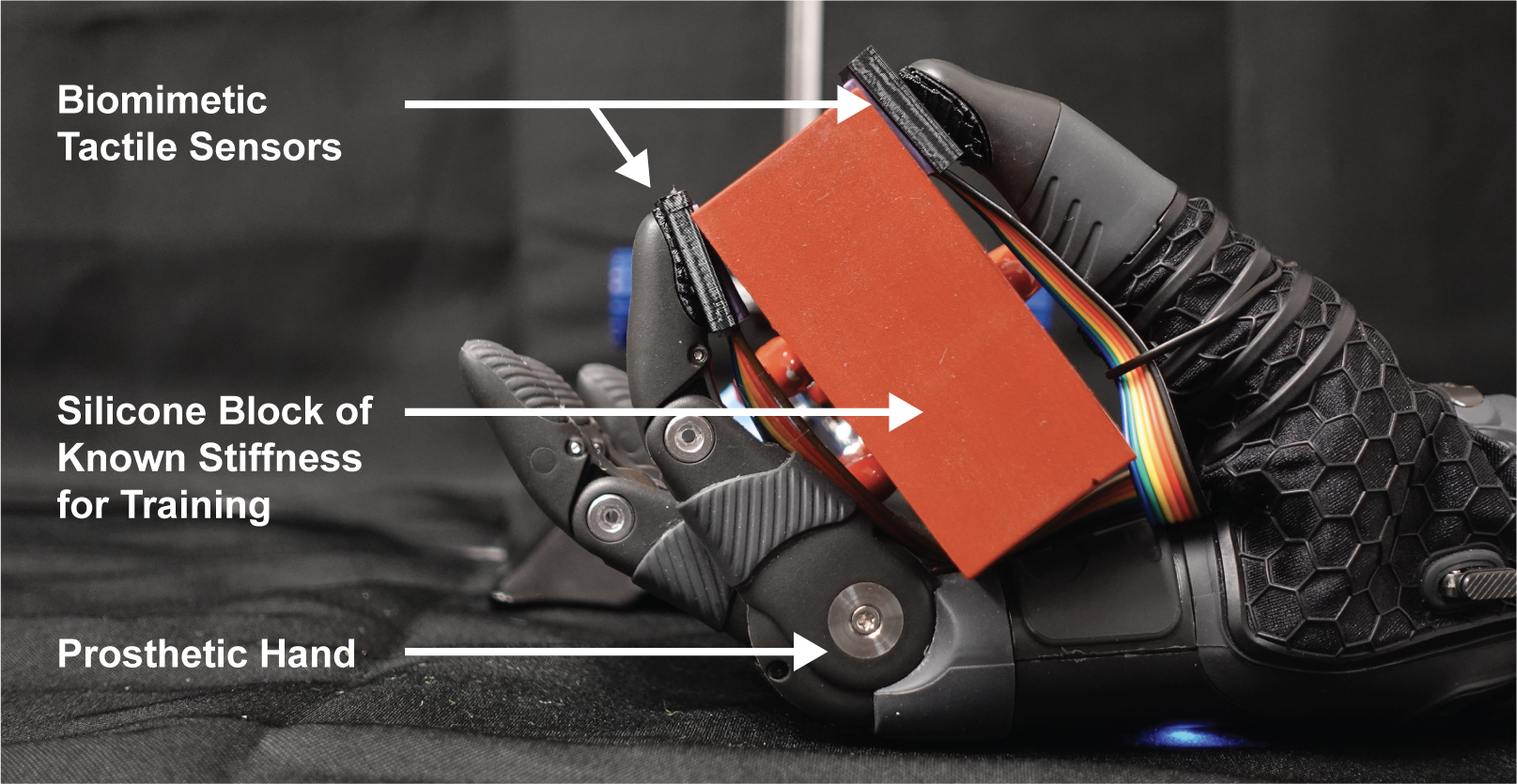}
    \caption{Training data collection with a prosthetic hand equipped with biomimetic tactile sensors, performing pinch tasks on silicone blocks.}
    \label{fig:TASKACombined}
\end{figure}

The prosthetic hand is positioned palm-up on a table, with silicone blocks suspended above in its grasp using a vertical clamp, as shown in Fig. \ref{fig:TASKACombined}. To train the models, 500 pinches per stiffness level (2,500 total) were performed, with slight adjustments to block placement to introduce natural variations and enhance dataset robustness.

\subsection{Contact Detection}

We implemented two approaches to detect the moment of contact from piezoelectric vibration data: a thresholding method and a machine learning-based method using a Support Vector Machine (SVM) with a Radial Basis Function (RBF) kernel. The thresholding method identified contact by applying a fixed threshold at three times the standard deviation (3$\sigma$) of steady-state signals \cite{Sahoo1988}. Both approaches operated on 20 ms sliding windows, including 17 ms of preceding data and 3 ms of newly acquired data. For ground truth validation, thin copper strips were applied to the finger and object surface and connected to a microcontroller's digital input pin to precisely register contact events. This timing was independently validated with LEDs connected to the contact strips on both fingers, recorded at 960 fps using a Sony RX100V camera (Sony Inc., Japan).

\subsection{Stiffness Discrimination and Regression Models}

To evaluate whether vibrational information at first contact encodes object stiffness, we implemented two approaches capable of extracting this information: a Support Vector Machine (SVM) and a Convolutional Neural Network (CNN), based on previous approaches with tactile data \cite{Gandarias2019, Alameh2020, Sankar2022}. Both models were applied in discrimination and regression contexts, trained on 2,500 pinches of known stiffness blocks, using a 90-10 training-validation split.

An SVM with RBF kernels was trained on 15 ms windows of vibrational data captured immediately after contact, leveraging its ability to model non-linear relationships between vibrational features and stiffness levels. Hyperparameter optimization was performed to fine-tune kernel parameters, maximizing discrimination accuracy and regression performance while maintaining computational efficiency.

The CNN approach utilized the EfficientNetV2 architecture, chosen for its efficiency in capturing complex data patterns \cite{Tan2021}. Inputs consisted of 15 ms temporal windows immediately after contact, identical in size to those used for the SVM. Training was performed using the Adam optimizer with an initial learning rate of 0.001. The Mean Squared Error (MSE) loss function was used for optimization, and training spanned 40 epochs. To ensure stability and convergence, a Step Learning Rate Scheduler reduced the learning rate by a factor of 0.5 every five epochs.

\subsection{Model Validation and Generalization}

To assess the generalizability of stiffness estimation, we tested them on objects representing real-world scenarios where a prosthetic hand must interact with materials of varying stiffness. The selected objects included two apples (28, 26 Shore A), two oranges (35, 37 Shore A), two tennis balls (45, 46 Shore A), and two avocados (59, 67 Shore A), as shown in Fig. \ref{fig:TASKACombined} (b) through (e). This selection covered a range of stiffness levels and material properties, reflecting the variability of everyday tasks and the challenges of adaptive manipulation. These objects were tested exclusively in a regression context to simulate real-world encounters with unknown objects, rather than evaluating similarities with pre-characterized ones through discrimination. Vibrational data from the initial 15 ms after contact served as input for the SVM and CNN models, which predicted stiffness values on a continuous scale. The model outputs were evaluated against ground truth, measured with the same Shore A durometer used for the training blocks, using a MSE metric.

\begin{figure}[t]
    \centering    
    \includegraphics[width=\linewidth]{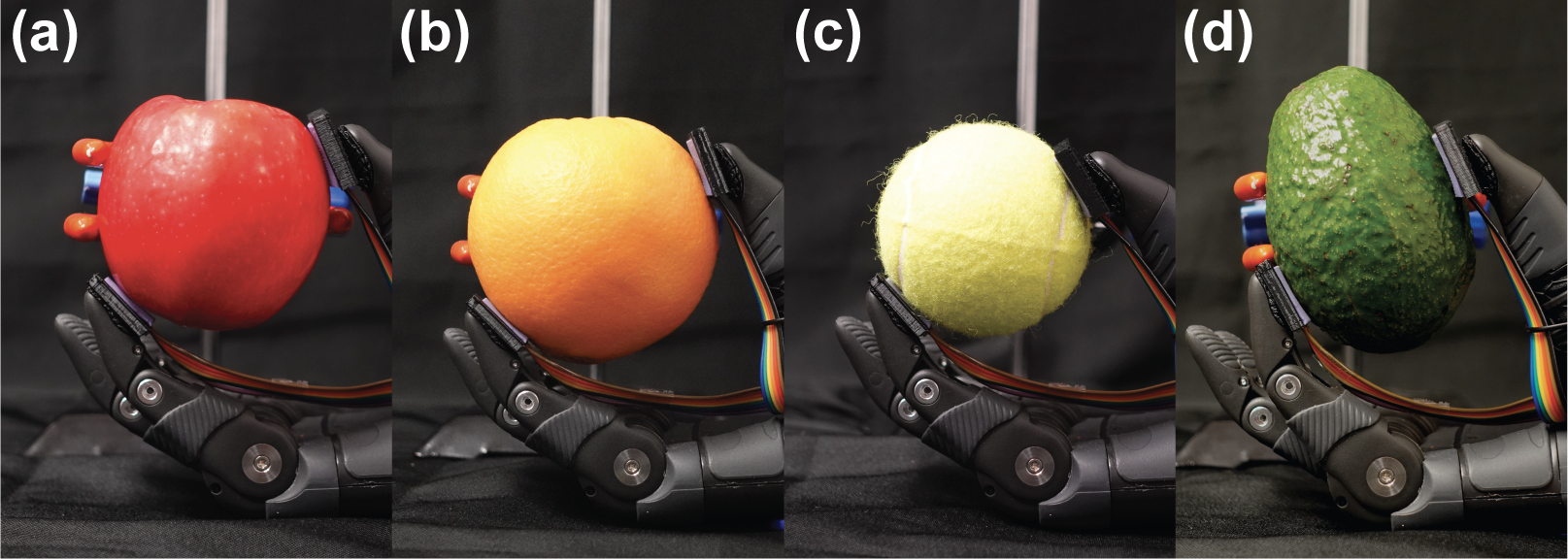}
    \caption{Validating regression models on real-world objects of diverse stiffness: (a) apples, (b) oranges, (c) tennis balls, and (d) avocados.}
    \label{fig:TASKAFruits}
\end{figure}

\section{Results and Discussion}


\begin{figure}[b!]
    \centering    
    \includegraphics[width=\linewidth]{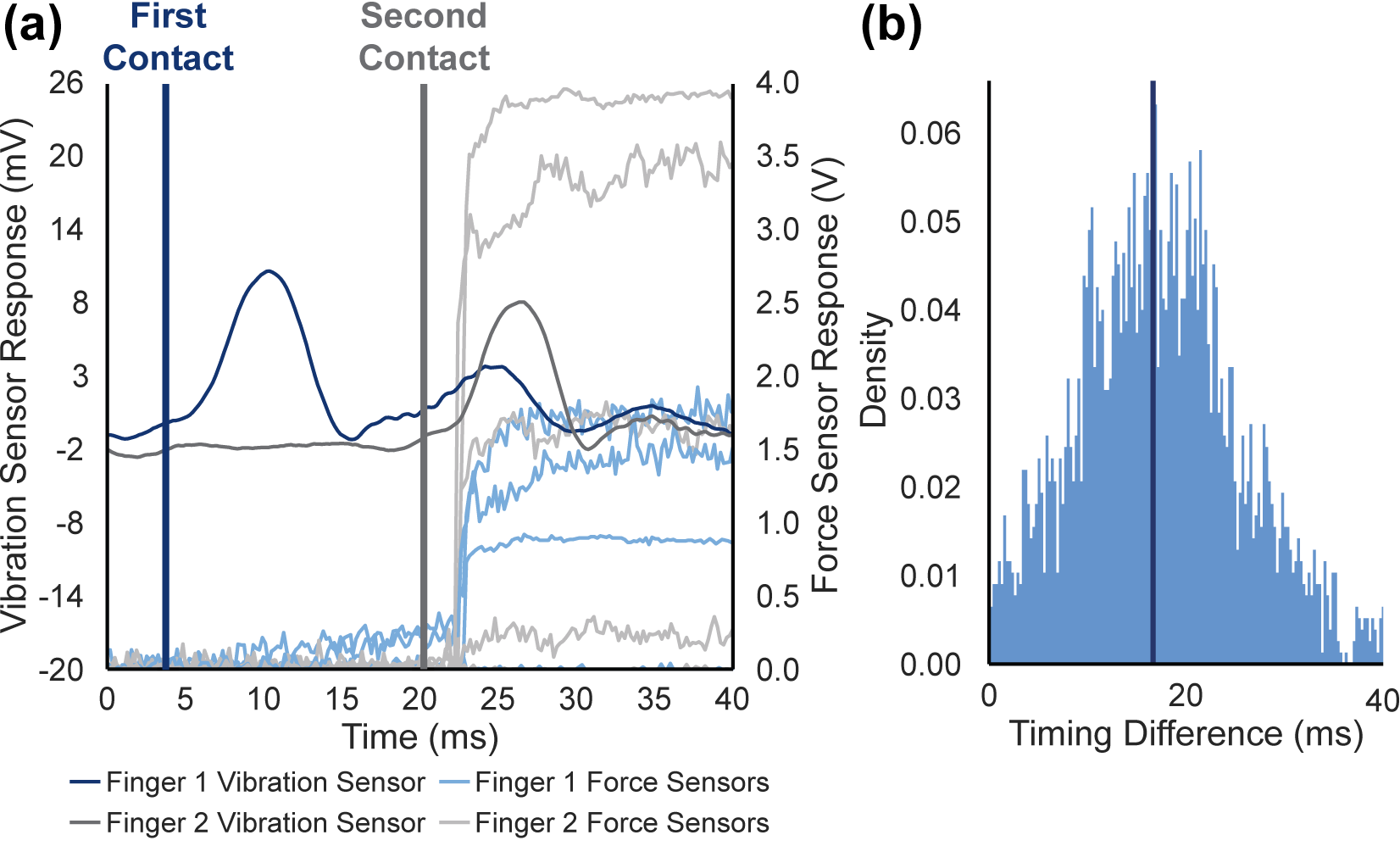}
    \caption{(a) Timing of contacts and corresponding sensor outputs for both fingers, and (b) distribution of time differences between first and second finger contacts.}
    \label{fig:TimingData}
\end{figure}

The contact event during the pinch grasp was detected using both thresholding and SVM methods. Thresholding achieved 90.6\% accuracy, hindered by false positives, while the SVM achieved a perfect 100\%. Despite its computational efficiency, thresholding's limitations led to the selection of the SVM for subsequent stiffness estimation tasks.

\begin{figure}[t!]
    \centering    
    \includegraphics[width=\linewidth]{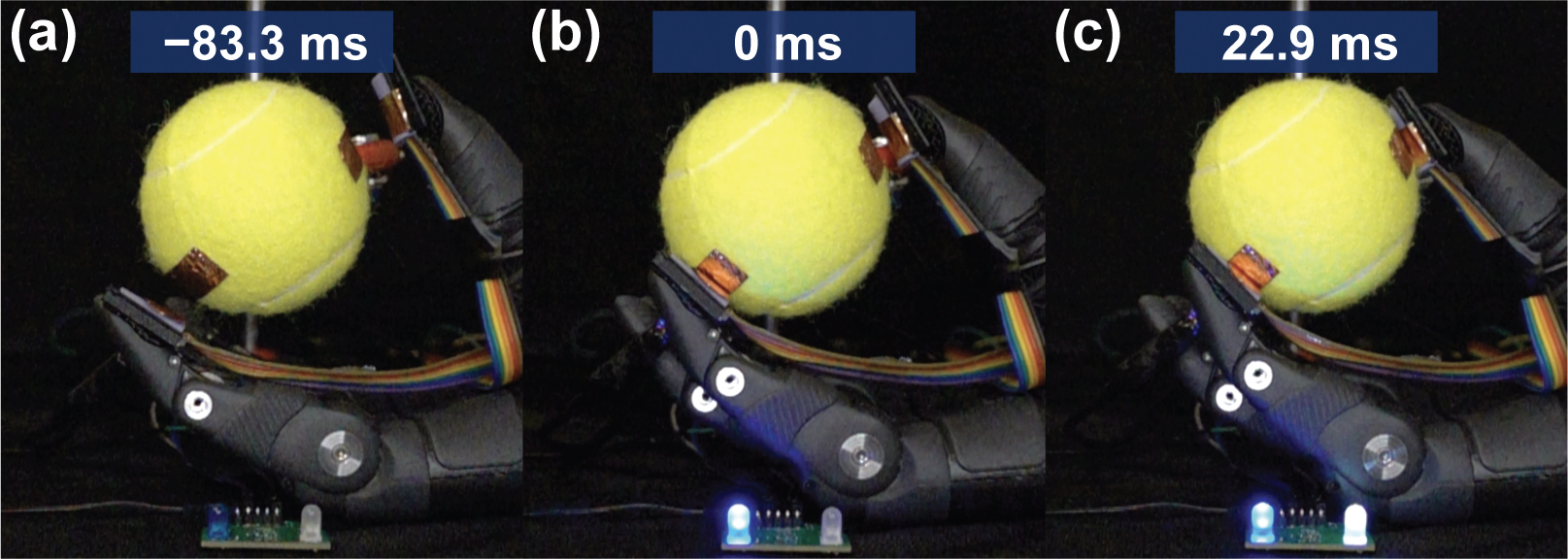}
    \caption{Contact events validation using 960 fps videography: (a) frame before contact, (b) frame at first contact, (c) frame at second contact.}
    \label{fig:SlowMo}
\end{figure}

The analysis of the pinch grasp event, where both fingers sequentially contact the object, reveals a measurable timing discrepancy between the two fingers. 
This \textit{asymmetry in finger contact}, confirmed by ground truth data from electrical contact strips as seen in Fig. \ref{fig:TimingData} (a), stems from the inherent unevenness of grasp dynamics, with one finger reaching the object before the other \cite{Schettino2013, Patel2017}. This stochastic phenomenon was quantified for our dataset, yielding a timing distribution with a mean offset of 16.65 ms and a standard deviation of 10.35 ms, as shown in Fig. \ref{fig:TimingData} (b). This timing difference was further validated through slow-motion videography, as shown in Fig. \ref{fig:SlowMo}. During the initial contact by the first finger, a pronounced vibrational signal is recorded, while force data remains absent due to the lack of a reaction force until the second finger establishes contact. This temporal gap provides a critical window to capture vibrational information before force is registered, enabling early data processing for real-time adaptive grasp modulation.

\begin{figure}[b!]
    \centering    
    \includegraphics[width=\linewidth]{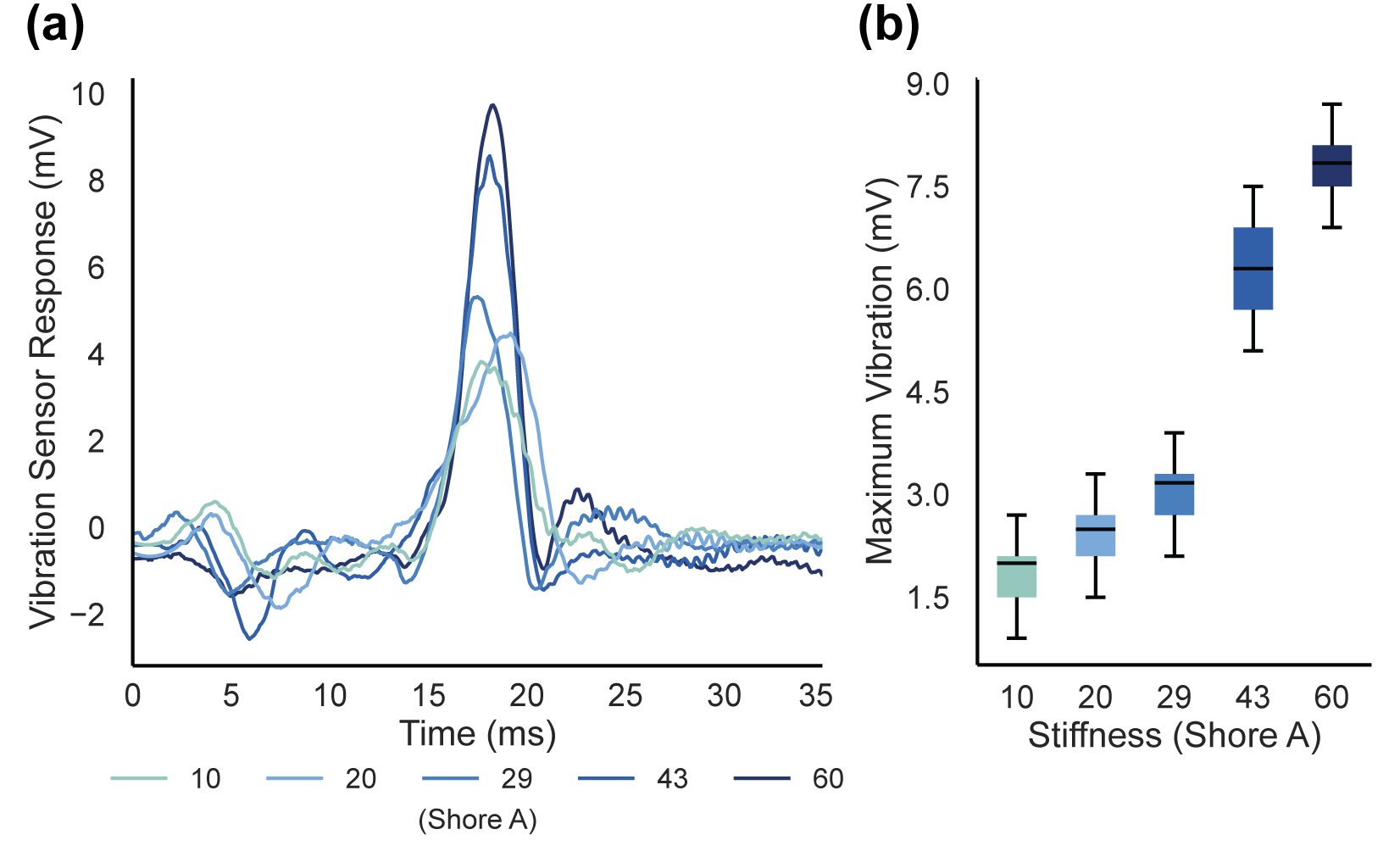}
    \caption{(a) Representative piezoelectric sensor responses for varying stiffness levels, and (b) maximum piezoelectric responses corresponding to different stiffness.}
    \label{fig:StiffnessData}
\end{figure}

The alignment of contact events across all stiffness blocks revealed an emergent relationship between the piezoelectric disk's response and the stiffness of the block, as seen in Fig. \ref{fig:StiffnessData} (a). Specifically, the amplitude of the piezoelectric response exhibited a proportionality to the stiffness as seen in Fig. \ref{fig:StiffnessData} (b). This suggests that the amplitude of the piezoelectric signal can reliably represent stiffness.

The SVM approach achieved an overall accuracy of 97.2\% on the discrimination task, while the CNN demonstrated a slightly higher accuracy of 98.6\%. While both models proved highly effective, the CNN's superior performance stems from its ability to learn and extract complex, high-dimensional features, capturing subtle vibrational patterns associated with stiffness \cite{Abiodun2019}. The SVM demonstrates superior computational efficiency, with an average inference time of less than 1 ms compared to the CNN's 1.4 ms. Both inference times are significantly shorter than the mean time between contacts (16.65 ms), making either approach viable for real-time stiffness estimation prior to fully grasping the object --- a critical advantage given the stochastic nature of grasp dynamics. This capability is unique, as previous approaches to stiffness estimation require a complete grasp of the object before any measurements or predictions can be made. This first contact approach can guide the development of faster, less intrusive methods for stiffness measurement. Additional optimization techniques can further enhance performance when these methods are deployed on embedded systems \cite{Baciu2024}.

\begin{figure}[t!]
    \centering    
    \includegraphics[width=\linewidth]{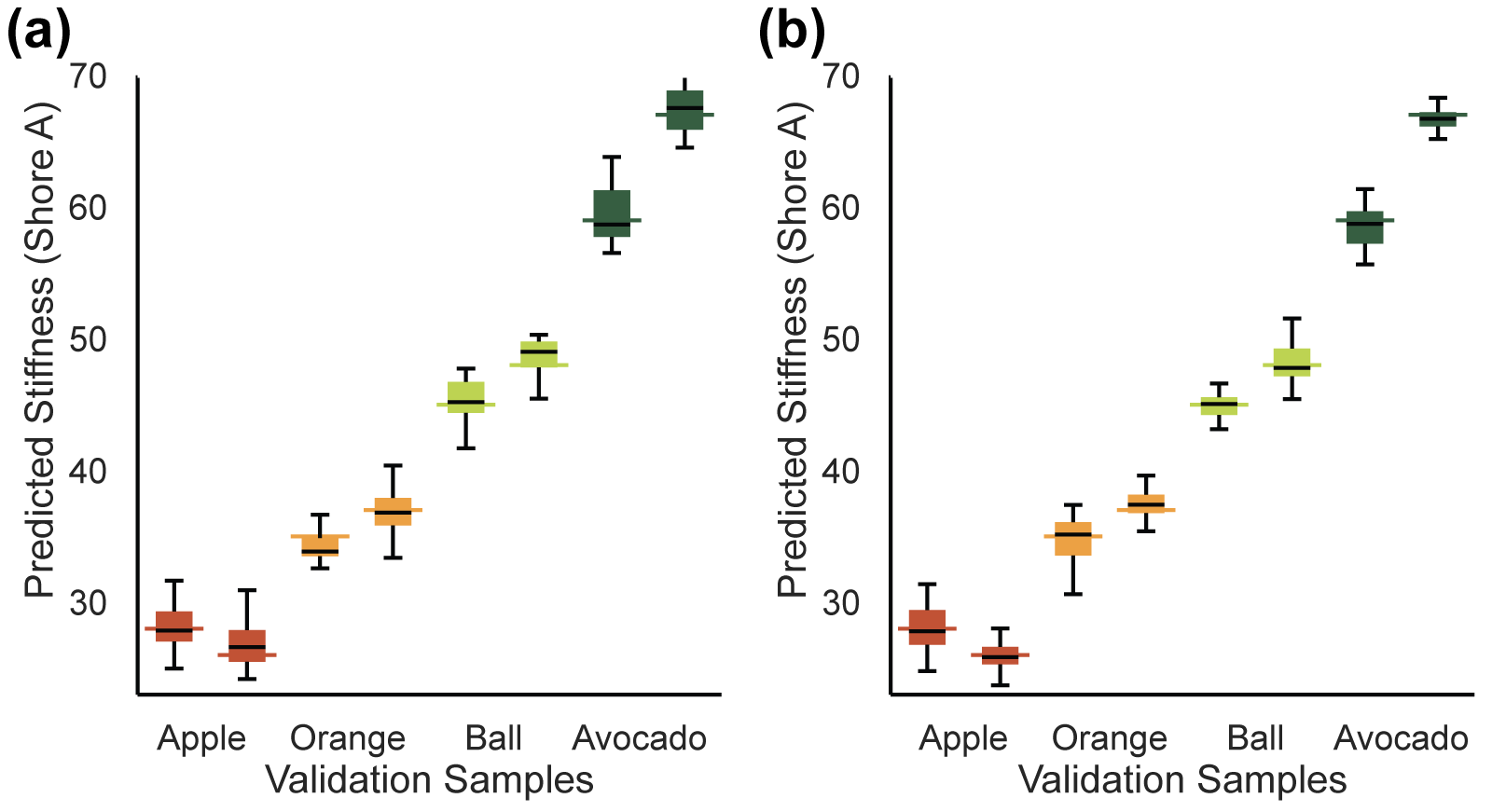}
    \caption{Stiffness regression on real-world objects with varying stiffness levels using (a) SVM and (b) CNN models, with true values represented by extended color bars alongside the box plots.}
    \label{fig:ValidationRegression}
\end{figure}

Both SVM and CNN approaches demonstrated strong generalization capabilities for stiffness estimation in objects with varying stiffness levels used in real-world settings. The models achieved overall MSE values of 4.17 ($<$1 ms inference time) and 2.39 ($<$1.5 ms inference time) Shore A, respectively, across all objects, as shown in Fig. \ref{fig:ValidationRegression}. These results and comparable latencies demonstrate that first-contact vibrational data generalizes effectively to unseen objects and can be utilized across diverse environments.

While the current setup was effective for controlled evaluations, expanding the dataset to include a wider range of objects and interaction scenarios will be crucial for developing more robust models. Future work should focus on integrating the proposed methods directly into prosthetic hand embedded systems, enabling true real-time grasp modulation and reducing communication delays between the hand and external microcontrollers. Other advanced prosthetics, such as the Vincent Hand (Vincent Systems GmbH, Germany), i-Limb Quantum (Össur, Iceland), and Bebionic Hand (Ottobock, Germany), could also benefit from incorporating this stiffness estimation method, enhancing control, responsiveness, and overall functionality \cite{Salminger2020, Guo2024}.

\section{Conclusion}

This study demonstrates the effectiveness of piezoelectric sensing for real-time stiffness estimation during the critical first contact phase of manipulation tasks. Vibrational data captured within 15 ms enabled accurate stiffness discrimination and regression, with inference times of $<$1 ms for the SVM and $<$1.5 ms for the CNN. These speeds, significantly faster than the 16.65 ms mean inter-contact interval, allow stiffness estimation before completely grasping the object --- a temporal advantage over prior methods reliant on complete grasp. The proportional relationship between piezoelectric response amplitude and stiffness confirms vibrational data as a reliable stiffness indicator. Integrating this method into embedded systems could enable real-time adaptive grasp modulation in prosthetics.

\section*{Acknowledgment}

\addcontentsline{toc}{section}{Acknowledgment}

\scriptsize
The authors would like to thank Shehreen Hassan and Shubhan Mathur for their contributions to the initial development and characterization of this work.

\normalsize

\bibliographystyle{IEEEtran}
\bibliography{paper.bib}

\end{document}